\documentclass[10pt,twocolumn,letterpaper]{article}

\usepackage[pagenumbers]{cvpr} 

\definecolor{cvprblue}{rgb}{0.21,0.49,0.74}
\usepackage[pagebackref,breaklinks,colorlinks,allcolors=cvprblue]{hyperref}

\usepackage{graphicx}
\usepackage{makecell}
\usepackage{multicol}
\usepackage{multirow}
\usepackage{amssymb}
\usepackage{pifont}
\usepackage{booktabs}
\usepackage{bbding}
\usepackage{amsmath}
\usepackage{algorithm}
\usepackage{algorithmic}
\usepackage{tabularx}
\usepackage{wrapfig}
\usepackage{xcolor}
\usepackage{relsize}

\def\confName{CVPR}
\def\confYear{2025}

\title{RFSR: Improving ISR Diffusion Models via Reward Feedback Learning}

\author{Xiaopeng Sun$^{1, \dagger}$, Qinwei Lin$^{1,2, \dagger}$\thanks{Work done during an internship at Meituan.}, Yu Gao$^{1, \dagger}$, Yujie Zhong$^{1}\footnotemark[3]$, Chengjian Feng$^{1}$,\\ Dengjie Li$^{1}$, Zheng Zhao$^{1}$, Jie Hu$^{1}$, Lin Ma$^{1}$\\
{\small $^1$Meituan Inc., $^2$Tsinghua University}\\
}

\begin{document}
\maketitle

\renewcommand{\thefootnote}{\fnsymbol{footnote}}
\footnotetext[2]{Equal contribution.}
\footnotetext[3]{Corresponding Author.}

\begin{abstract}

Generative diffusion models (DM) have been extensively utilized in image super-resolution (ISR). Most of the existing methods adopt the denoising loss from DDPMs for model optimization. We posit that introducing reward feedback learning to finetune the existing models can further improve the quality of the generated images. In this paper, we propose a timestep-aware training strategy with reward feedback learning. Specifically, in the initial denoising stages of ISR diffusion, we apply low-frequency constraints to super-resolution (SR) images to maintain structural stability. In the later denoising stages, we use reward feedback learning to improve the perceptual and aesthetic quality of the SR images. In addition, we incorporate Gram-KL regularization to alleviate stylization caused by reward hacking. Our method can be integrated into any diffusion-based ISR model in a plug-and-play manner. Experiments show that ISR diffusion models, when fine-tuned with our method, significantly improve the perceptual and aesthetic quality of SR images, achieving excellent subjective results. Code: \url{https://github.com/sxpro/RFSR}

\end{abstract}
    
\section{Introduction}
\label{sec:intro}

Recently, diffusion models emerge as a powerful alternative for image generation and restoration tasks. Denoising diffusion probabilistic models (DDPMs)~\cite{ho2020denoising, rombach2022high, saharia2022photorealistic} demonstrate exceptional performance in approximating complex distributions, making them suitable for various image processing applications, including image super-resolution (ISR). Unlike generative adversarial networks, diffusion models exploit strong image priors and can generate high-quality images by progressively refining noisy inputs. This capability is extended to real-world ISR scenarios. Recent approaches begin to exploit their potential to address the real-world ISR challenge. StableSR~\cite{wang2024exploiting} and DiffBIR~\cite{lin2023diffbir} rely solely on the input low-resolution image as a control signal. PASD~\cite{yang2023pasd} directly uses standard high-level models to effectively extract semantic cues. SeeSR~\cite{wu2024seesr} aligns the captions of LR and ground truth (GT) images, then incorporates the captions as an additional control condition for the text-to-image (T2I) model. These diffusion-based ISR methods are all fine-tuned based on pre-trained stable diffusion models, and primarily use the denoising loss from the DDPMs for model optimization.

Large Language Models (LLMs) and Text-to-Image (T2I)
models\cite{dong2023raft, lee2023aligning, yang2024dense, yuan2024self, zhang2024confronting, ren2025byteedit, li2025controlnet} experience a significant surge in incorporating learning based on human feedback, achieving outstanding performance across various benchmarks and subjective evaluations. Inspired by these developments, we aim to introduce reward feedback learning into ISR diffusion models by employing both subjective and objective reward models to improve ISR performance. 

\begin{figure*}[!th]
  \centering
  \begin{subfigure}{0.5\linewidth}
    \includegraphics[width=\linewidth, trim={0 0 0 0}, clip]{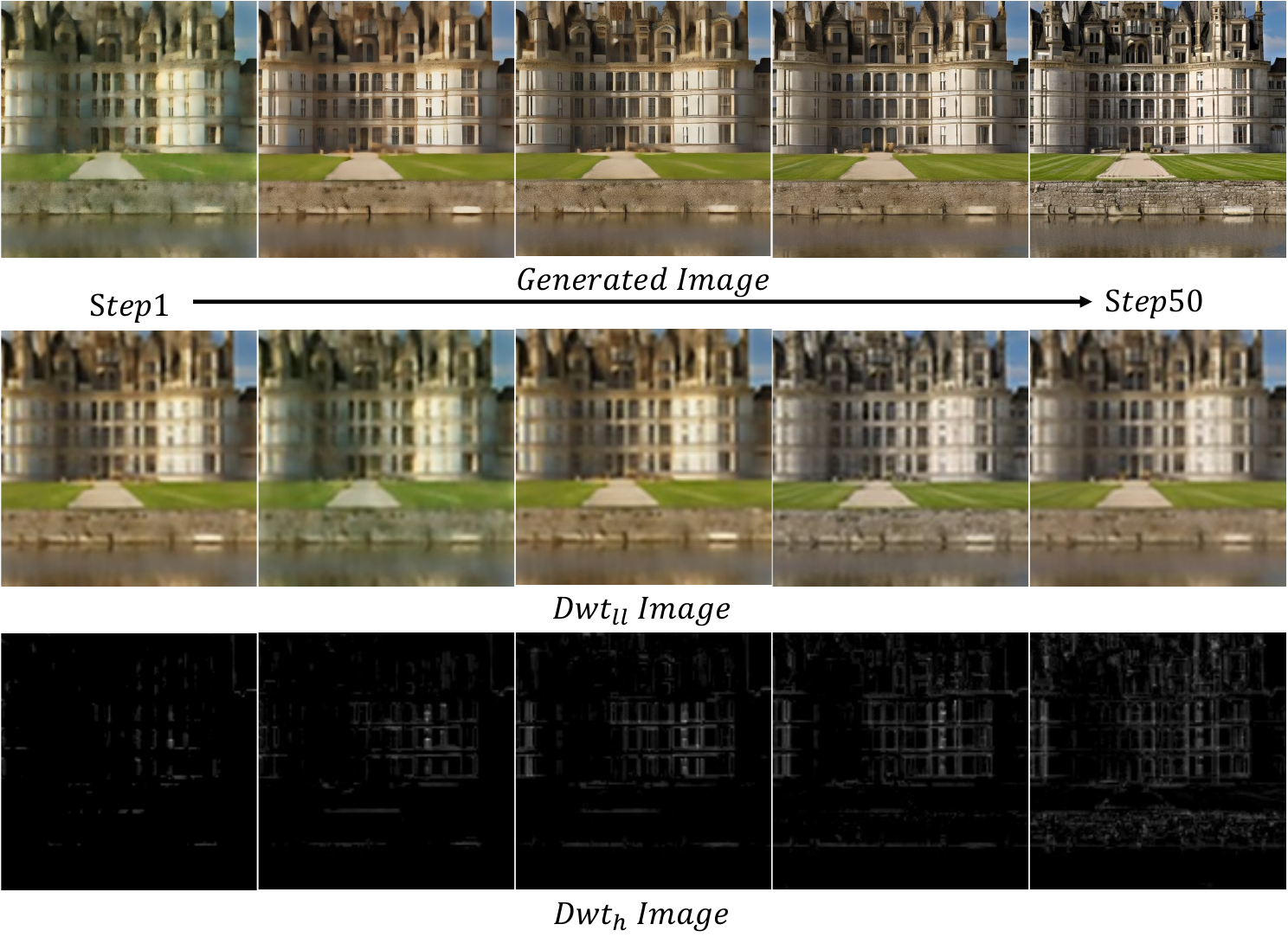}
    \caption{The denoising process.}
    \label{fig:short-a}
  \end{subfigure}
  \hfill
  \begin{subfigure}{0.45\linewidth}
    \includegraphics[width=\linewidth, trim={0 0 0 0}, clip]{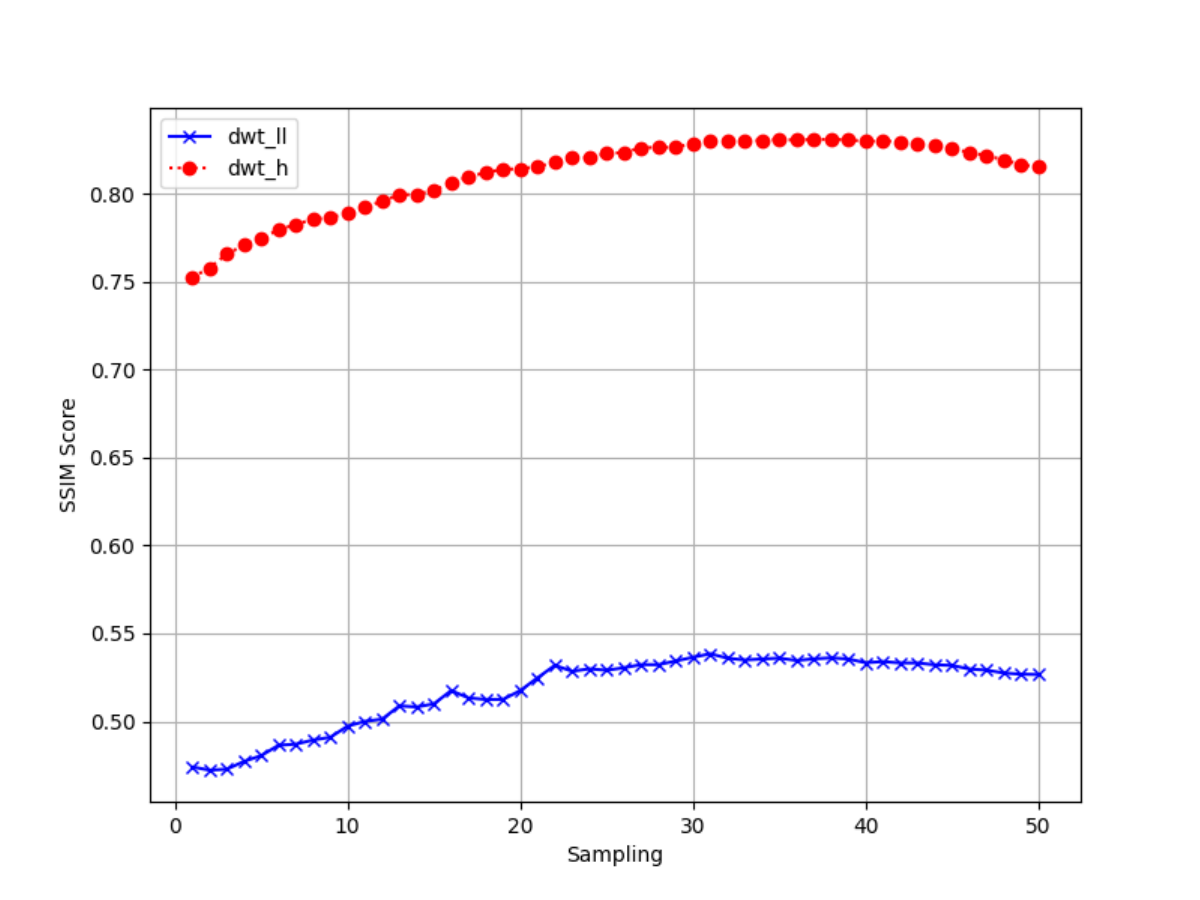}
    \caption{The correlation between SeeSR and ground truth in the frequency domain on DIV2K-val with respect to SSIM and sampling steps.}
    \label{fig:short-b}
  \end{subfigure}
  \caption{(a) The first row shows the progressive denoising of the image during the iterative process, while the next two rows show the low-frequency and high-frequency components derived from each stage of the DWT transformation. Clearly, once the low-frequency components reach stability, their fluctuations decrease, while the high-frequency components become increasingly complex. (b) At smaller steps, both the low-frequency and high-frequency information are close to the ground truth. As the number of steps increases, these frequency components gradually diverge from the GT. This observation leads us to maintain the structural stability of the SR images in the early steps and to encourage the ISR diffusion model to generate more perceptually pleasing and detailed texture information in later steps.
}
  \label{fig:Motivation}
\end{figure*}

Specifically, we analyze the denoising process of the ISR diffusion model using SeeSR as an example, which uses 50 steps of DDIM sampling during inference. As shown in Figure~\ref{fig:Motivation}, unlike many T2I methods~\cite{si2024freeu, xu2024imagereward} based on reward feedback learning, the ISR diffusion model already has a relatively complete contour at the sampling step (\(st\)). 1. This observation inspired us to apply supervision to the ISR diffusion model at step 1 (i.e., when \( T = 1000 \)).
During each step of the SeeSR denoising process, we decode the latent noise into super-resolution (SR) images using a VAE decoder~\cite{rombach2022high}. We then compare the SSIM~\cite{wang2004image} values of each intermediate SR image with the ground truth image in both low and high frequency bands, computed using the Discrete Wavelet Transform (DWT). We do not use PSNR because SSIM incorporates a normalization process that helps to minimize the effect of image distribution within the dataset. It is evident that the SSIM values for the low frequency information in the SR images consistently improve during the initial phases. This implies that the ISR diffusion model emphasizes the reconstruction of low frequency information that encapsulates the basic structure of the image. In the later steps, high frequency information gradually accumulates. However, after 40 steps, the SSIM score begins to decrease and the high frequency details of the SR images increasingly diverge from those of the ground truth. This suggests that the diffusion process generates more complex and unrestricted high frequency information. 

\begin{figure*}
  \centering
  \includegraphics[width=\linewidth]{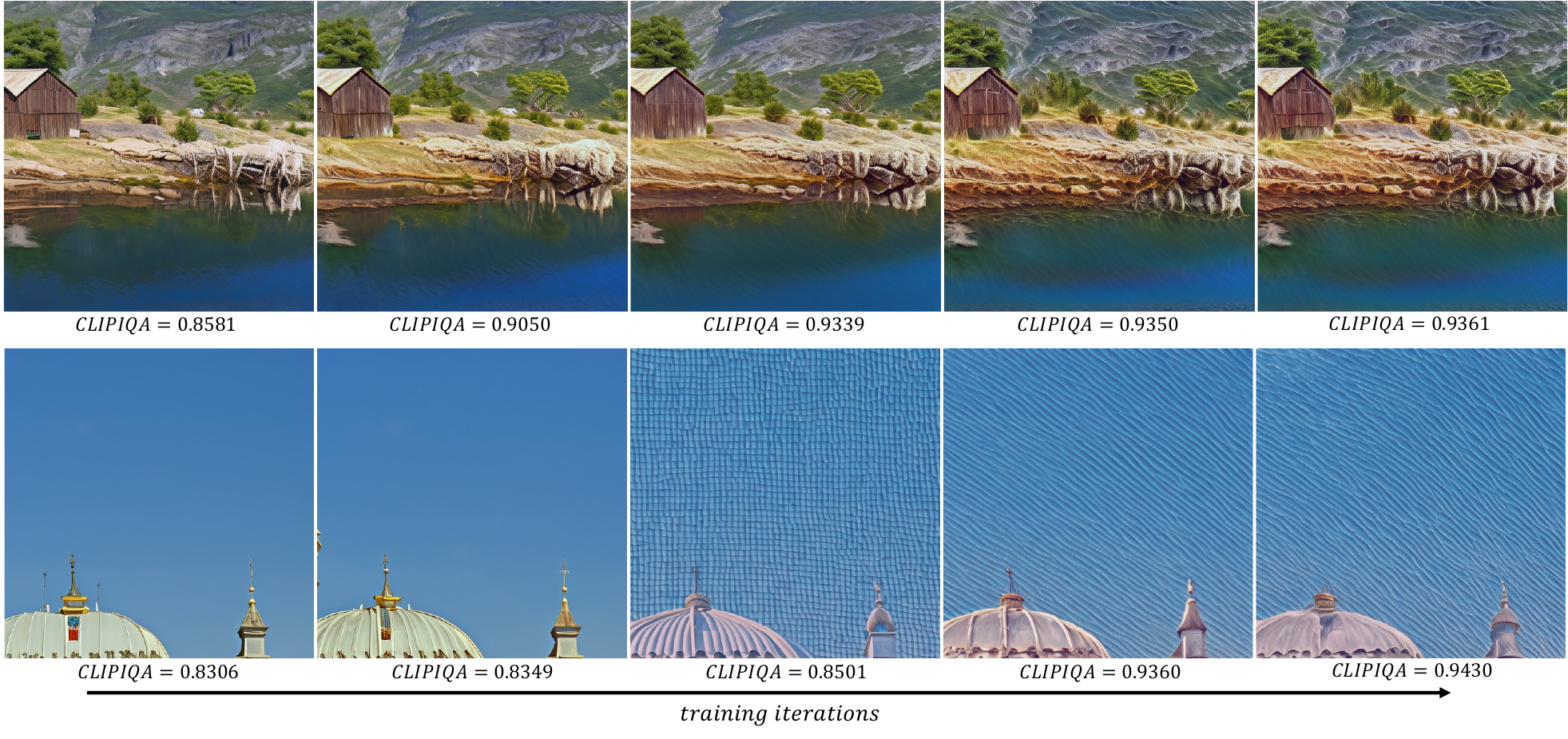}
  \caption{{Visualization for Reward Hacking.} The direct application of reward feedback learning significantly improves the perceptual metrics (e.g., CLIPIQA) of SR images, but leads to reward hacking, resulting in progressively degrading image quality. The subjective manifestation of this issue is that SR images tend to adopt a specific stylization and generate strange lines.}

  \label{fig:reward_hacking}
\end{figure*}

Based on these observations, we use different rewards at different steps to incentivize the diffusion model. Specifically, in the early denoising steps, we use the low-frequency information from the ground truth to constrain the generated images, thereby improving image fidelity. In the later denoising steps, we use subjective quality rewards to motivate the diffusion model to improve perceptual and aesthetic image quality. Since the direct use of subjective quality rewards can lead to image stylization due to reward hacking~\cite{clark2023directly}, as shown in Figure~\ref{fig:reward_hacking}, increased training leads to higher CLIPIQA~\cite{yang2022maniqa} scores but worse subjective results. Therefore, we apply Gram-KL regularization at this stage to alleviate the stylization effects.

Overall, the main contributions are summarized as follows:
\begin{itemize}
    \item [$\bullet$] We are the first to introduce reward feedback learning into super-resolution fine-tuning, paving new ways to improve model performance.
    \vspace{1mm}
    \item [$\bullet$] We introduce a timestep-aware training approach to drive reward feedback learning. Specifically, during the initial denoising steps, we impose constraints on the low-frequency information of SR images to maintain structural stability. In the later denoising steps, we employ reward models to improve the subjective generation quality of the ISR models.
    \vspace{1mm}
    \item [$\bullet$] We propose a method for regularizing the Gram matrix of the generated images to alleviate image stylization issues caused by reward hacking. 
    \vspace{1mm}
    \item [$\bullet$] Our proposed method can be integrated into any diffusion-based ISR model in a plug-and-play manner. Extensive experimental results demonstrate the effectiveness of this method in improving the fine-tuning of diffusion-based ISR models, showing significant improvements in image clarity, detail preservation, and overall perceptual quality.
\end{itemize}

\section{Related Work}
\label{sec:related}

\subsection{Diffusion-based Image Super-Resolution}

Recent approaches have exploited the implicit knowledge in pre-trained diffusion models by using large-scale text-to-image (T2I) diffusion models trained on large high-resolution image datasets. These models provide enhanced capabilities for processing diverse content. StableSR~\cite{wang2024exploiting}is a pioneering work that fine-tunes the Stable Diffusion (SD)~\cite{stabilityai2023} model by training a time-aware encoder and employs feature warping to balance fidelity and perceptual quality, thereby improving fidelity by utilizing prior information from diffusion models. On the other hand, DiffBIR~\cite{lin2023diffbir} combines traditional pixel regression-based image recovery with text-to-image diffusion models. PASD~\cite{yang2023pasd} directly uses high-level models to effectively extract semantic cues. SeeSR~\cite{wu2024seesr} incorporates DAPE to align the captions of LR and GT images. XPSR\cite{qu2025xpsr} introduces more negative prompts to improve the SR performance of the model.

\subsection{Reward Feedback Learning}

Xu~\cite{xu2024imagereward} use reward function gradients to fine-tune diffusion models. They evaluate the reward of a predicted clean image at a randomly selected step t in the denoising trajectory, rather than evaluating it on the final image. Generally, any perceptual model that takes images as input and makes predictions can function as a reward model. Commonly used reward models for fine-tuning text-to-image diffusion models include CLIP scores for text-image alignment~\cite{clark2023directly, kim2022diffusionclip, radford2021learning}, human preferences\cite{kirstain2023pick, clark2023directly, prabhudesai2023aligning}, and JPEG\cite{black2023training, clark2023directly} compressibility.
In this study, we explore the use of timestep-aware reward feedback learning to fine-tune ISR diffusion models, introducing Gram-KL regularization to alleviate the phenomenon of reward hacking.

\section{Method}
\label{sec:method}

\subsection{Motivation}
By visualizing the inference process of the super-resolution diffusion model in Figure~\ref{fig:Motivation}, we observe that as the number of sampling steps increases, the model first reconstructs the overall structure of the LR image and then adds texture details. Therefore, we aim to impose low-frequency information constraints in the early stages of the diffusion model's denoising process and apply reward feedback learning for high-frequency information in the later stages.

\subsection{Low-Frequency Structure Constraint}

The low-frequency information in an image often contains the overall structure of the image content. Compared to GAN-based super-resolution~\cite{wang2021real}, diffusion-based super-resolution models have stronger generative capabilities. However, they are more likely to produce structures that do not match the input image. Therefore, by constraining the low-frequency information of the generated image in the early stages of the diffusion process, we can better maintain the consistency of the image structure without affecting the generation of texture details.

In this section, we extract the low-frequency information of both the ground truth ($I_{gt}$) image and the generated image based on the Discrete Wavelet Transform (DWT). Given an image \(I_t \in \mathbb{R}^{H \times W \times 3}\) obtained at time \(t\) by the VAE decoder, we use DWT to extract its low-frequency components, which contain the overall structure and coarse details of the image. We define:
\begin{equation}
\text{DWT}(\cdot) : \mathbb{R}^{H \times W \times 3} \rightarrow \mathbb{R}^{4 \times \frac{H}{2} \times \frac{W}{2} \times 3},
\label{eq:dwt}
\end{equation}
which contains one low-frequency image and three high-frequency images. Since we only need the low-frequency image, we use \(\text{DWT}(I_t)_{LL}\). Therefore, the constraint for the low-frequency information can be defined as follows:

\begin{equation}
\mathcal{L}_{dwt_{ll}} = \left| \text{DWT}(I_{\text{gt}})_{LL} - \text{DWT}(I_t)_{LL} \right|.
\label{eq:dwtabs}
\end{equation}

\begin{figure*}[!ht]
  \centering
  \includegraphics[width=\linewidth]{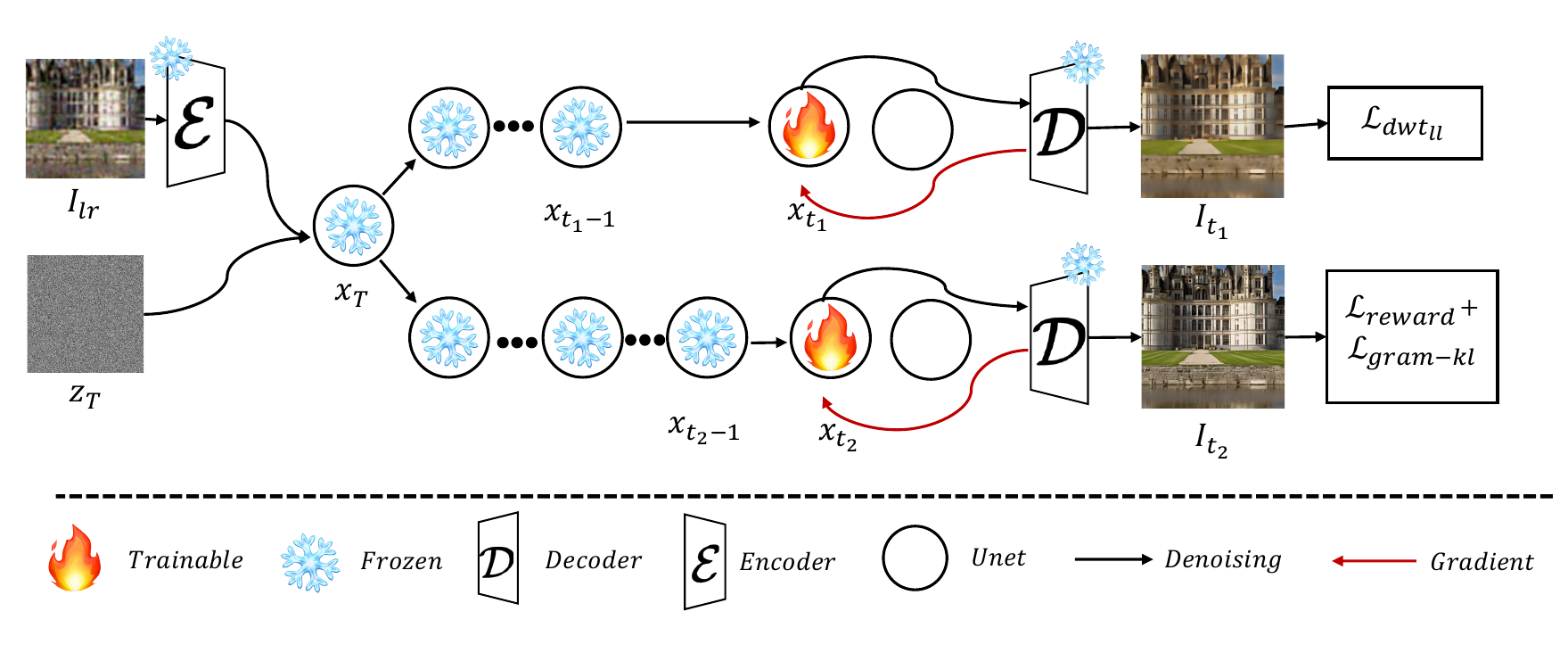}
  \caption{{Overview of our method.}}
  \label{fig:overview}
\end{figure*}

\subsection{Reward Feedback Learning}

To significantly improve the subjective performance of the super-resolution model, we introduce reward feedback learning to fine-tune the parameters $\theta$ in the super-resolution model $G$. Unlike most diffusion methods, which refine predictions sequentially from the last step $x_T$ to the initial step $x'_0$ ($x_T \rightarrow x_{T-1} \rightarrow \cdots \rightarrow x'_0$), we adopt an innovative approach by optimizing the prediction results of intermediate time steps $t \in [0, T]$ starting from $T$ ($x_T \rightarrow x_{T-1} \rightarrow \cdots \rightarrow x_t$). Specifically, we define the image at time step $t$ starting from $T$ as follows:

\begin{equation}
I_t = G_{\theta}(z_t, I_{lr}, t, c_{v}, c_{t}),
\label{eq:xt}
\end{equation}
where $I_{lr}$ is the LR image, $z_t$ is the noisy latent, $c_{v}$ is the condition from ControlNet~\cite{zhang2023adding}, and $c_{t}$ is the text embedding. Note that DiffBIR does not include $I_{lr}$ and $c_{t}$ in the process from $x_T$ to $x_0$. 

Therefore, our optimization objective is to minimize the loss of the reward model (RM) at time step $t$, which is 
\begin{equation}
\begin{aligned}
\mathcal{L}_{reward} &= \mathcal{L}(RW(c_t, I_t)) \\
                  &= \mathcal{L}(RM(c_t, G_{\theta}(z_t, I_{lr}, t, c_{v}, c_{t}))).
\label{eq:reward}
\end{aligned}
\end{equation}
\noindent \textbf{Reward Feedback Models.}
To improve the subjective quality of ISR, we choose CLIP-IQA~\cite{yang2022maniqa} and Image Reward (IW)~\cite{xu2024imagereward} as our RW models. CLIP-IQA is a method based on the Contrastive Language–Image Pre-training (CLIP) model, which is used to evaluate the quality and feel of images. Through CLIP-IQA, our approach improves the perceptual quality of SR images. IW is a model designed to learn and evaluate human preferences for text-to-image generation. Through IW, our method enables ISR to generate images that are more aligned with human preferences. Therefore, the reward loss function is as follows:
\begin{equation}
\begin{aligned}
\mathcal{L}_{{reward}} &= \mathcal{L}(RW(c_t, I_t)) \\
                  &= \lambda_{clipiqa}\mathcal{L}_{\text{CLIP-IQA}}(I_t) + \lambda_{iw}\mathcal{L}_{\text{IW}}(c_t, I_t),
\label{eq:reward_loss}
\end{aligned}
\end{equation}
where $\lambda_{clipiqa}$ and $\lambda_{iw}$ are hyperparameters. The RW is a versatile model that can be selected, such as models trained on different datasets to capture human preferences. Different RW models will provide different benefits, which are beyond the scope of this paper.

\begin{table*}[!ht]
    \centering
    \renewcommand\arraystretch{1.3}
    \setlength{\tabcolsep}{1mm}{
    \begin{tabular}{c|c|cc|cc|cc|cc}
    \toprule
    {Datasets} & {Metrics} 
    & DiffBIR & DiffBIR-RFSR & DiffBIR-tag & \makecell{DiffBIR-\\tag-RFSR} & PASD
    & PASD-RFSR & SeeSR   &SeeSR-RFSR    
     \\ \midrule
    \multirow{5}{*}{DIV2K-val} 
    
    &$\text{MANIQA}\uparrow$ 
    &0.4869 &\textbf{0.5058} &0.5193
    &\textbf{0.5341} &0.3412 &\textbf{0.4517} &0.5091 &\textbf{0.5954}
    \\
    &$\text{MUSIQ}\uparrow$ 
    &67.88 &\textbf{69.2538} &69.27 
    &\textbf{70.2611} &50.26 &\textbf{59.88} &67.40 &\textbf{69.97} 
    \\
    &$\text{CLIPIQA}\uparrow$ 
    &0.7036 &\textbf{0.7231} &0.7114 
    &\textbf{0.7377} &0.4619 &\textbf{0.6056} &0.6989 &\textbf{0.7944} 
    \\   
    &$\text{Aesthetic}\uparrow$ 
    &5.0447 &\textbf{5.0973} &5.1826
    &\textbf{5.2521} &4.7717 &\textbf{5.0758} &5.1475 &\textbf{5.2683 }
    \\ 
    &$\text{LPIPS}\downarrow$ 
    &\textbf{0.3659} &0.3882 &\textbf{0.3556}
    &0.3693 &0.4410 &\textbf{0.4277} &\textbf{0.3329} &0.3369 
    \\
    
    \midrule
    \multirow{5}{*}{DRealSR} 
    
    &$\text{MANIQA}\uparrow$ 
    &0.4923 &\textbf{0.4978} &0.5229
    &\textbf{0.5388} &0.3688 &\textbf{0.5130}&0.5146 &\textbf{0.5922}
    \\
    &$\text{MUSIQ}\uparrow$ 
    &65.73 &\textbf{66.2777} &67.44
    &\textbf{68.7545} &50.18 &\textbf{63.79} &64.92 &\textbf{67.48}
    \\
    &$\text{CLIPIQA}\uparrow$ 
    &\textbf{0.6842} &{0.6830} &0.7038 
    &\textbf{0.7174} &0.4872 &\textbf{0.6708} &0.6813 &\textbf{0.7596}
    \\   
    &$\text{Aesthetic}\uparrow$ 
    &4.6101 &\textbf{4.6184} &4.7418
    &\textbf{4.8171} &4.4174 &\textbf{4.6433} &4.6985 &\textbf{4.8158 }
    \\   
    &$\text{LPIPS}\downarrow$ 
    &\textbf{0.3497} &0.3933 &\textbf{0.3480}
    &0.3729 &\textbf{0.2413} &0.2654 &\textbf{0.2346} &0.2761
    \\
    \midrule
        
    \multirow{5}{*}{RealSR} 
    
    &$\text{MANIQA}\uparrow$ 
    &\textbf{0.4857} &{0.4805} &0.5365
    &\textbf{0.5431} &0.3941 &\textbf{0.5316} &0.5428 &\textbf{0.6057}
    \\
    &$\text{MUSIQ}\uparrow$ 
    &68.19 &\textbf{68.507} &70.04 
    &\textbf{70.9008} &59.83 &\textbf{68.99} &69.77 &\textbf{71.22}
    \\
    &$\text{CLIPIQA}\uparrow$ 
    &\textbf{0.6897} &{0.6862} &0.7048 
    &\textbf{0.7155} &0.4788 &\textbf{0.6491} &0.6611 &\textbf{0.7438}
    \\   
    &$\text{Aesthetic}\uparrow$ 
    &4.8153 &\textbf{4.8299} &4.8964
    &\textbf{4.9212} &4.675 &\textbf{4.7879 }&4.8046 &\textbf{4.8985 }
    \\   
    &$\text{LPIPS}\downarrow$ 
    &\textbf{0.2760} &0.2838 &\textbf{0.2969}
    &0.3153 &\textbf{0.2252} &0.2468 &\textbf{0.2354} &0.2642 
    \\
    \bottomrule
    \end{tabular}}
    \caption{Quantitative comparison results are presented on both synthetic and real-world benchmark datasets. For each of the comparison groups, better results are highlighted in bold. The $\downarrow$ indicates that the smaller values are better, while the $\uparrow$ indicates that the larger values are better.}
    \label{tab:quantitative_comparsion}
\end{table*}

\subsection{Alleviating Reward Hacking with Gram-KL}
Directly employing reward models as loss functions can lead to reward hacking issues \cite{clark2023directly}, where the perceptual metrics remain very high as the number of training iterations increase, but the actual visual quality deteriorates, as shown in Figure~\ref{fig:reward_hacking}.
Previous work \cite{clark2023directly} employs LoRA and early termination strategies, as well as latent noise regularization constraints \cite{fan2024reinforcement}. However, these methods generally have constraint objectives that are inconsistent with the model optimization objectives, often resulting in a trade-off. In super-resolution, we observe that such hacking phenomena often manifest as strong stylization. 

Based on this issue, we propose a stylization regularization constraint, which is orthogonal to the generation objectives, thereby further mitigating the hacking phenomenon. We use KL divergence to regularize the Gram matrices~\cite{gatys2016image} of the super-resolution images between the training model G and the pretrained model G', as follows:
\begin{equation}
\begin{aligned}
\mathcal{L}_{gram-kl} = \left\| Gram(Vgg(G_{\theta}(z_t, I_{lr}, t, c_{v}, c_{t}))) - \right. \\
\left. Gram(Vgg(G_{\theta'}(z_t, I_{lr}, t, c_{v}, c_{t}))) \right\|_2^2,
\end{aligned}
\label{eq:gram_kl}
\end{equation}
where $Vgg$ is a classic and widely used feature extractor~\cite{simonyan2014very}, $Gram$ refers to the Gram matrix computed from feature maps, which represents the style of an image. 
We do not use the gram matrix of the GT here because current pre-trained ISR diffusion models do not exhibit reward hacking, and we aim to generate images with quality that surpasses the GT.

\subsection{Timestep-aware Training}

As previously discussed, when the time step \(t_1\) is relatively large or the sampling step \(st_1\) is relatively small (i.e., \(t_1 \in [600, 1000]\) or \(st_1 \in [1, 20]\), representing the first 40\%), we optimize the model using \(\mathcal{L}_{{dwt_{ll}}}\). Conversely, if the time step \(t_2\) is relatively small or the sampling step \(st_2\) is relatively large (i.e., \(t_2 \in [0, 200]\) or \(st_2 \in [41, 50]\), representing the last 20\%), we optimize the model using \(\mathcal{L}_{{reward}} + \mathcal{L}_{{gram-kl}}\). Therefore, the loss function is defined as follows:
\begin{equation}
\mathcal{L}oss =
\begin{cases} 
    \lambda_{dwt}\mathcal{L}_{dwt_{ll}}, & \text{if } 
    \begin{aligned} 
        t &\in [T, t_{1}]\\

    \end{aligned} \\[1.5em]
    \mathcal{L}_{\text{reward}} + \lambda_{r}\mathcal{L}_{\text{gram-kl}}, & \text{if } 
    \begin{aligned} 
        t &\in [0, t_{2}] \\ 

    \end{aligned}
\end{cases}
\label{eq:piecewise_function}
\end{equation}

where \( t \in [T, t_{1}] \) is equivalent to \( st \in [1, st_{1}] \), and \( t \in [0, t_{2}] \) is equivalent to \( st \in [st_{2}, st_{\text{latest}}] \), and $\lambda_{r}$ and $\lambda_{dwt}$ are hyperparameters. In particular, if gradient updates are enabled during the entire process from \(T\) to 0 during training, it can lead to gradient explosion. Therefore, we enable gradient updates only in the final step. According to the studies by~\cite{clark2023directly, xu2024imagereward}, enabling gradient updates in the final step also provides a certain fine-tuning effect. The detailed training process is shown in Figure~\ref{fig:overview} and the Supplementary Material.

\section{Experiments}
\label{sec:exp}

\begin{figure*}[!ht]
  \centering
  \includegraphics[width=\linewidth]{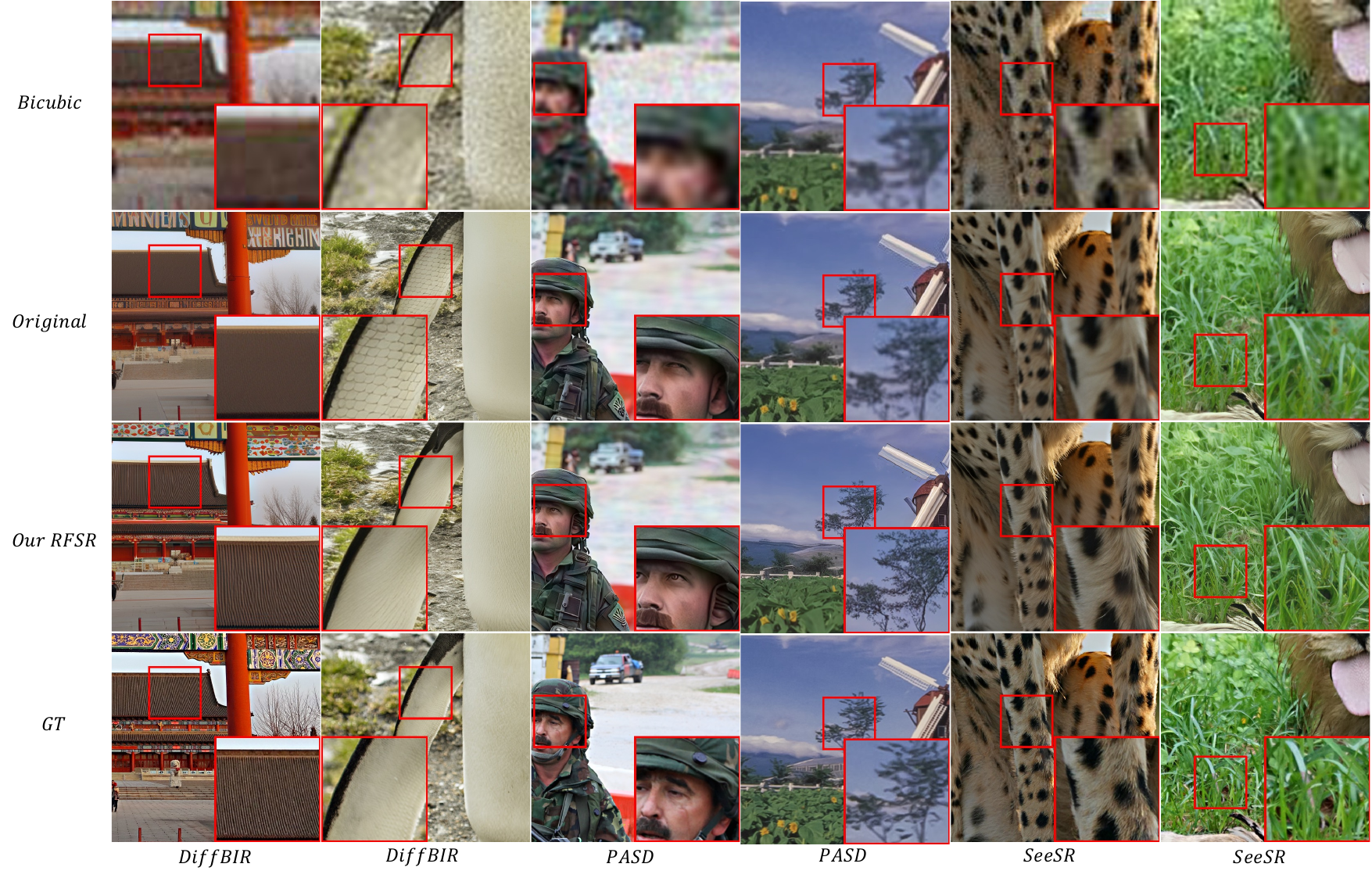}
  \caption{{A visual comparison of state-of-the-art ISR diffusion models and their counterparts trained with our RFSR is presented. Each row, from top to bottom, displays the results of bicubic interpolation, the original ISR model, the ISR model trained with our RFSR, and the GT image. Please zoom in for a better view.}}
  \label{fig:Qualitative}
\end{figure*}

Our approach is plug-and-play for diffusion-based ISR, so we select some representative and state-of-the-art works for our experiments: DiffBIR~\cite{lin2023diffbir} (using the v1 model from the DiffBIR paper), SeeSR~\cite{wu2024seesr}, and PASD~\cite{yang2023pasd} (using the MSCOCO I2T model from the PASD project).

\subsection{Implementation Details}

We fine-tune these models using the Adam~\cite{kingma2014adam} optimizer with a learning rate of $5 \times 10^{-6}$ and a batch size of 8, with the ground truth image resolution set to 512$\times$512. The training is conducted for 10,000 iterations on two A100-80G GPUs, with gradients enabled only for the U-Net and ControlNet. The inference steps follow the settings of each method: for DiffBIR and SeeSR, $st_{1}$ is set to 20, $st_{2}$ is set to 40, and $st_{latest}$ is set to 50 (consistent with their respective papers). For PASD, $st_{1}$ is set to 8, $st_{2}$ is set to 17, and $st_{latest}$ is set to 20 (consistent with their respective papers). We set $\lambda_{dwt}$ to 0.0005, $\lambda_{clipiqa}$ to 0.00005, and both $\lambda_{iw}$ and $\lambda_{r}$ to 0.000005. Furthermore, to ensure the stability of model parameter updates, we introduce an Exponential Moving Average (EMA) decay parameter and set it to 0.999.

\subsection{Dataset and Evaluation Metric}
\noindent \textbf{Datasets.}
We fine-tune the models on DIV2K~\cite{agustsson2017ntire}, DIV8K~\cite{gu2019div8k}, Flickr2K~\cite{timofte2017ntire}, OST~\cite{wang2018recovering}, and the first 10K face images from FFHQ~\cite{karras2019ffhq}. The degradation pipeline of Real-ESRGAN~\cite{wang2021real} is used to synthesize low-resolution and high-resolution training pairs. We evaluate our approach on both synthetic and real-world datasets. The synthetic dataset is generated from the DIV2K validation set following the Real-ESRGAN degradation pipeline. For real-world test datasets, we use RealSR~\cite{cai2019toward} and DRealSR~\cite{wei2020component} for evaluation. 

In particular, since DiffBIR has no text input, the image reward model cannot fully implement reward feedback learning. Therefore, we use SeeSR's DAPE as the text encoder for DiffBIR. During training, tags obtained from low-resolution images through DAPE are fed into both DiffBIR and the image reward model. Thus, when evaluating the fine-tuning performance of DiffBIR, we compare the results with and without captions.

\noindent \textbf{Metrics.}
In order to comprehensively evaluate the performance of different methods, we employ a range of widely used reference and non-reference metrics. LPIPS~\cite{zhang2018unreasonable} is reference-based perceptual quality metric. MANIQA~\cite{yang2022maniqa}, MUSIQ~\cite{yang2022maniqa}, and CLIPIQA~\cite{yang2022maniqa} are non-reference image quality metrics. The aesthetic score~\cite{schuhmann2022laion} is used to evaluate the aesthetic quality of images and is trained to predict the aesthetic aspects of the generated images.

\subsection{Comparison of Diffusion-based ISR with RFSR}

\noindent \textbf{Quantitative Comparisons.}
We perform quantitative evaluations on the DIV2K-val, DRealSR, and RealSR datasets. As shown in Table~\ref{tab:quantitative_comparsion}, the methods fine-tuned with RFSR achieve significant improvements in both perceptual and subjective metrics. For example, on the DRealSR dataset, PASD-RFSR achieves maximum improvements of 39\% over the pre-trained MANIQA model, 37\% over CLIPIQA, 27\% over MUSIQ, and 5\% over Aesthetic. This demonstrates that our subjective reward feedback learning effectively improves the performance of the ISR diffusion model.

\vspace{1mm}
\noindent \textbf{Qualitative Comparisons.}
We provide visual comparisons in Figure~\ref{fig:Qualitative}. With a comprehensive understanding of the scene information and enhanced by RFSR, diffusion-based ISR excels at enhancing high-quality texture details. In the DiffBIR column, our method restores the textures that the original model loses or incorrectly reconstructs. In the PASD column, our method adeptly reconstructs realistic textures such as facial features and tree and plant characteristics. Similarly, as shown in the SeeSR column, our results show significantly clearer and more realistic features in animals. Conversely, better LPIPS does not necessarily lead to better subjective effects, as shown by DiffBIR. By keeping the loss of fidelity metrics within an acceptable range and subsequently improving the perceptual and aesthetic metrics, we achieve superior visual results.

\subsection{Ablation Study}
\label{sec:abl}
Among these ISR diffusion models, SeeSR exhibits superior overall capabilities, ensuring stable and high-quality super-resolution (SR) results across various scenarios. Consequently, we employ SeeSR-RFSR in our ablation experiments.

\begin{figure}[!h]
  \centering
  \includegraphics[width=\linewidth]{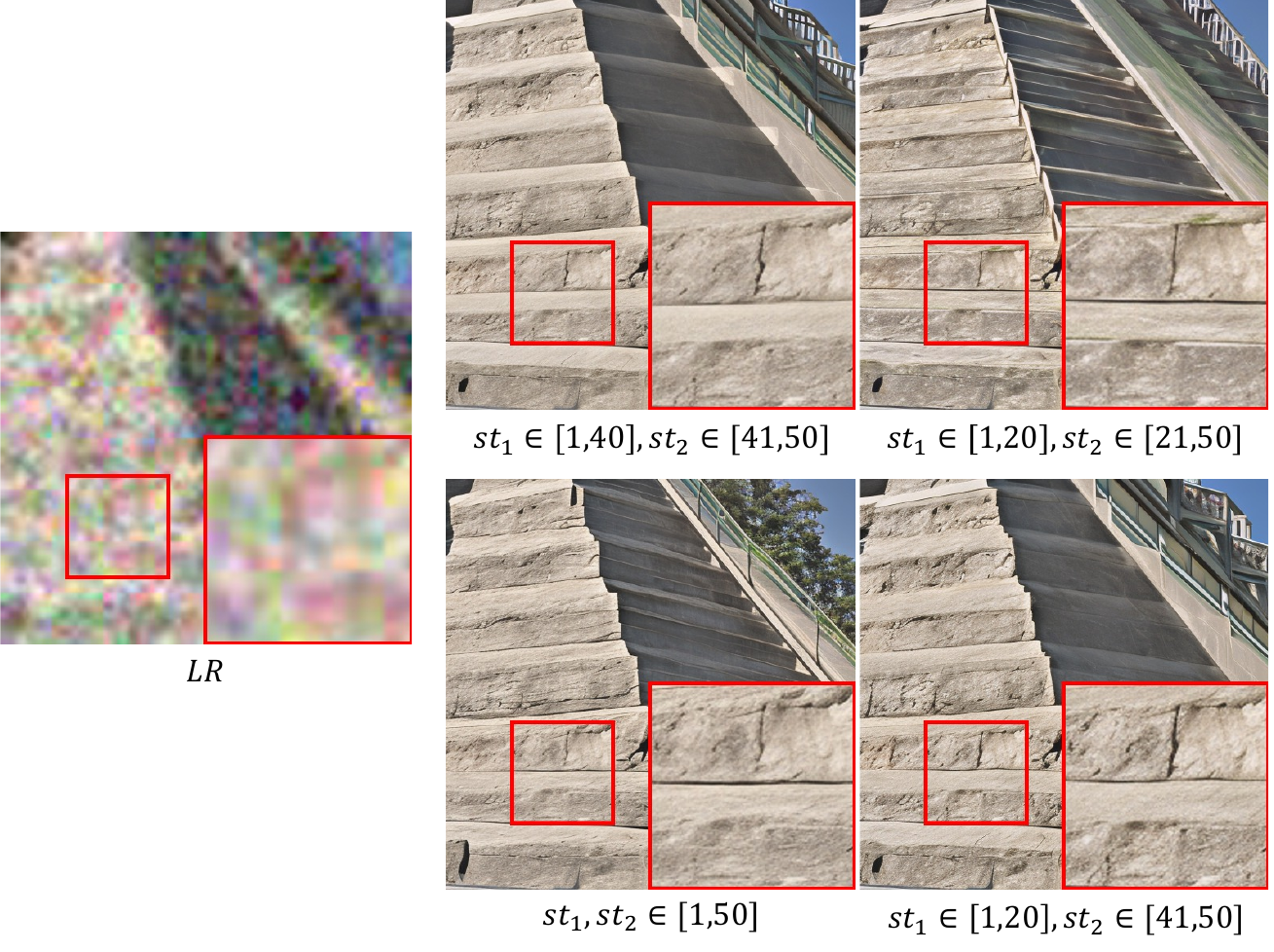}
  \caption{{Effectiveness of Timestep-Aware Training.} An excessively large $st_1$ interval causes image blurring, while an overly large $st_2$ interval induces image stylization.
}
  \label{fig:ablation_timestep}
\end{figure}

\vspace{1mm}
\noindent \textbf{Effectiveness of Timestep-aware Training.}
We adjust the intervals of $st_1$ and $st_2$, as presented in Table~\ref{tab:ablation_ts}. When we increase the interval length of $st_1$, it enhances the constraint on low-frequency information, which significantly improves fidelity metrics such as LPIPS. However, perceptual and aesthetic metrics experience considerable declines. As shown in Figure~\ref{fig:ablation_timestep}, increasing the $st_1$ interval too much results in blurred images. Similarly, widening the $st_2$ interval results in improved perceptual metrics but reduced image fidelity. Consequently, SR images exhibit reward hacking, as the stair sections begin to adopt an oil painting style, as shown in the figure.
Additionally, stylization caused by reward hacking appears. When we remove the intervals for $st_1$ and $st_2$—meaning that ISR diffusion applies identical constraints and rewards across all $st$—the model's metrics become mediocre, and the subjective effects exhibit stylization caused by reward hacking. Therefore, our time-aware strategy is highly effective, generating more realistic textures while maintaining corresponding image quality scores, thus achieving a good trade-off between subjective effects and objective metrics.

\begin{table}[!h]
    \centering
    \resizebox{0.48\textwidth}{!}{%
    \begin{tabular}{c ccccc} \toprule
      Experiments

      &$\text{LPIPS}\downarrow$ 
      &$\text{MANIQA}\uparrow$ 
      &$\text{MUSIQ}\uparrow$ 
      &$\text{CLIPIQA}\uparrow$ 
      &$\text{Aesthetic}\uparrow$ 
  \\ \midrule 
    $st_{1} \in [1, 40], \; st_{2} \in [41,50]$ & 0.3347 & 0.5660 & 69.81 & 0.7751 & 5.2499  \\
   
    $st_{1} \in [1, 20], \; st_{2} \in [21,50]$ & 0.3453 & 0.6044 & 71.29 & 0.8058 & 5.3024  \\
    $st_{1}, \; st_{2} \in [1,50]$ & 0.3389 & 0.5915 & 71.69 & 0.8030 & 5.3373  \\
     \midrule 
    Ours & 0.3369 & 0.5954 & 69.97 & 0.7944 & 5.2683  \\ \bottomrule
    \end{tabular}}
    \caption{{Ablations of Timestep-Aware Training.}}
    \label{tab:ablation_ts}
\end{table}

\begin{figure}[!h]
  \centering
  \includegraphics[
    width=\linewidth,
    trim=0 1.5cm 0 0, 
    clip
  ]{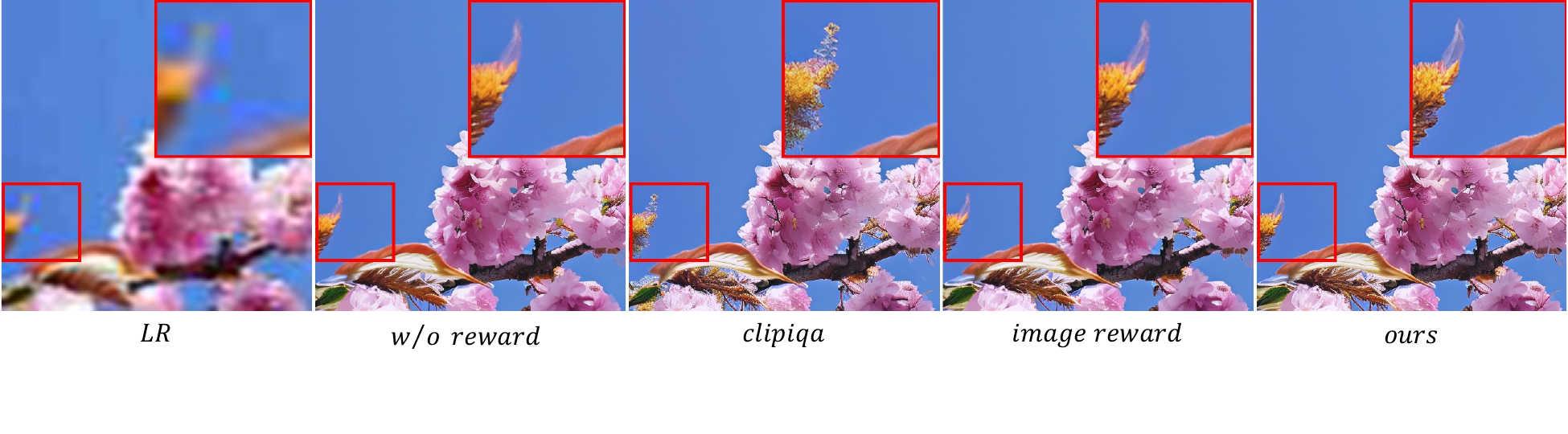}
  \caption{{Effectiveness of Reward Feedback Models.} The introduction of CLIPIQA enables ISR to generate more intricate details, while incorporating Image Reward allows ISR to generate more coherent and aesthetically pleasing textures.}
  \label{fig:ablation_reward}
\end{figure}

\vspace{1mm}
\noindent \textbf{Effectiveness of Reward Feedback Models.}
As illustrated in Table~\ref{tab:ablation_rf}, without reward models, only $\mathcal{L}_{dwt_{ll}}$ constrains the ISR model, resulting in relatively high fidelity. However, as depicted in Figure~\ref{fig:ablation_reward}, the subjective effects are extremely blurred. We conduct ablation studies on the reward models, revealing that when only CLIP-IQA is utilized as the reward, the ISR diffusion model achieves high perceptual metrics. However, SR images tend to be sharper yet are prone to generating incorrect textures and more noise. Conversely, when IW alone is used as the reward, the ISR diffusion model shows commendable subjective performance, with SR images being more coherent but less clear and less textured. Therefore, by incorporating both perceptual and aesthetic rewards into the reward models, we improve the clarity of SR images while maintaining aesthetic and appropriate texture structures within the images.

\begin{table}[!h]
    \centering
    \resizebox{0.45\textwidth}{!}{%
    \begin{tabular}{cc ccccc} \toprule
      $\mathcal{L}_{CLIP-IQA}$
      & $\mathcal{L}_{IW}$

      & $\text{LPIPS}\downarrow$ 
      & $\text{MANIQA}\uparrow$ 
      & $\text{MUSIQ}\uparrow$ 
      & $\text{CLIPIQA}\uparrow$ 
      & $\text{Aesthetic}\uparrow$ 
      
      \\ \midrule 

        \XSolidBrush   & \XSolidBrush & 0.3311 & 0.5206 & 68.08 & 0.7146 & 5.1667  \\	
        \XSolidBrush   & \Checkmark   & 0.3310 & 0.4698 & 66.35 & 0.6671 & 5.1426  \\
        \Checkmark    & \XSolidBrush  & 0.3373 & 0.5839 & 69.80 & 0.7934 & 5.2402  \\ 
        \midrule 
        \Checkmark    & \Checkmark   & 0.3369 & 0.5954 & 69.97 & 0.7944 & 5.2683 \\ \bottomrule
    \end{tabular}}
    \caption{{Ablations of Reward Feedback Models.}}
    \label{tab:ablation_rf}
\end{table}

\begin{figure}[!h]
  \centering
  \includegraphics[width=\linewidth]{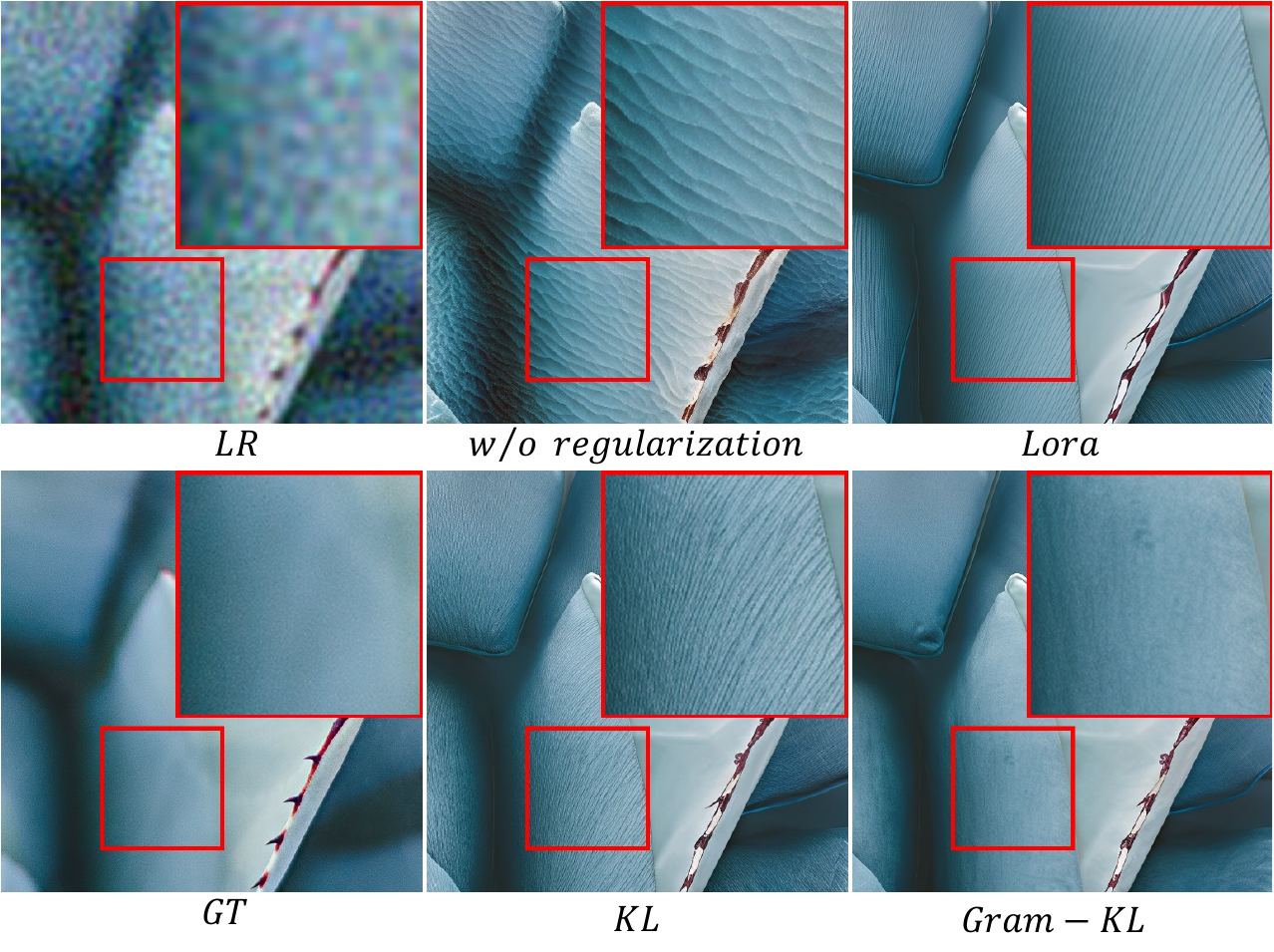}
  \caption{{Effectiveness of Style Regularization.}Without regularization, SR images exhibit high clarity but possess an oil-painting-like quality, generating strange lines. Similarly, LORA and KL result in pronounced image stylization and generate strange lines. In contrast, Gram-KL regularization preserves the natural style of the images, producing clearer results and richer textures.
}
  \label{fig:ablation_kl}
\end{figure}

\noindent \textbf{Effectiveness of Alleviating Image Stylization.}
We compare several methods for alleviating image stylization caused by reward hacking. LoRA is discussed in \cite{clark2023directly}, and KL is addressed in \cite{fan2024reinforcement}. Under the same training conditions, it is evident that our method is the most effective in alleviating stylization effects. As shown in Table~\ref{tab:ablation_kl}, although LoRA and KL achieve higher scores on perceptual metrics, as illustrated in Figure~\ref{fig:ablation_kl}, their regularization effects remain limited, resulting in more stylized outputs from the ISR diffusion model. While our Gram-KL produces similar subjective results, Gram-KL generates images with greater clarity and more distinct textures. Thus, Gram-KL effectively suppresses stylization while exploiting the generative capabilities of the diffusion process.

\begin{table}[!h]
    \centering
    \resizebox{0.45\textwidth}{!}{%
    \begin{tabular}{c ccccc} \toprule
      exps

      &$\text{LPIPS}\downarrow$ 
      &$\text{MANIQA}\uparrow$ 
      &$\text{MUSIQ}\uparrow$ 
      &$\text{CLIPIQA}\uparrow$ 
      &$\text{Aesthetic}\uparrow$ 
      
  \\ \midrule 
    w/o regularization  & 0.4062 & 0.6615 & 70.86 & 0.8964 & 5.3612   \\
    LoRA  & 0.3449 & 0.6171 & 71.20 & 0.7834 & 5.2669   \\
    KL & 0.3377 & 0.5908 & 69.89 & 0.8000 &  5.2719	  \\
     \midrule 
    Gram-KL(Ours)            & 0.3369 & 0.5954 & 69.97 & 0.7944 & 5.2683  \\ \bottomrule
    \end{tabular}}
    \caption{{Ablations of Alleviating Image Stylization.}}
    \label{tab:ablation_kl}
\end{table}

\begin{figure}[!h]
  \centering
  \includegraphics[
    width=\linewidth,
    trim=0 1.5cm 0 0, 
    clip
  ]{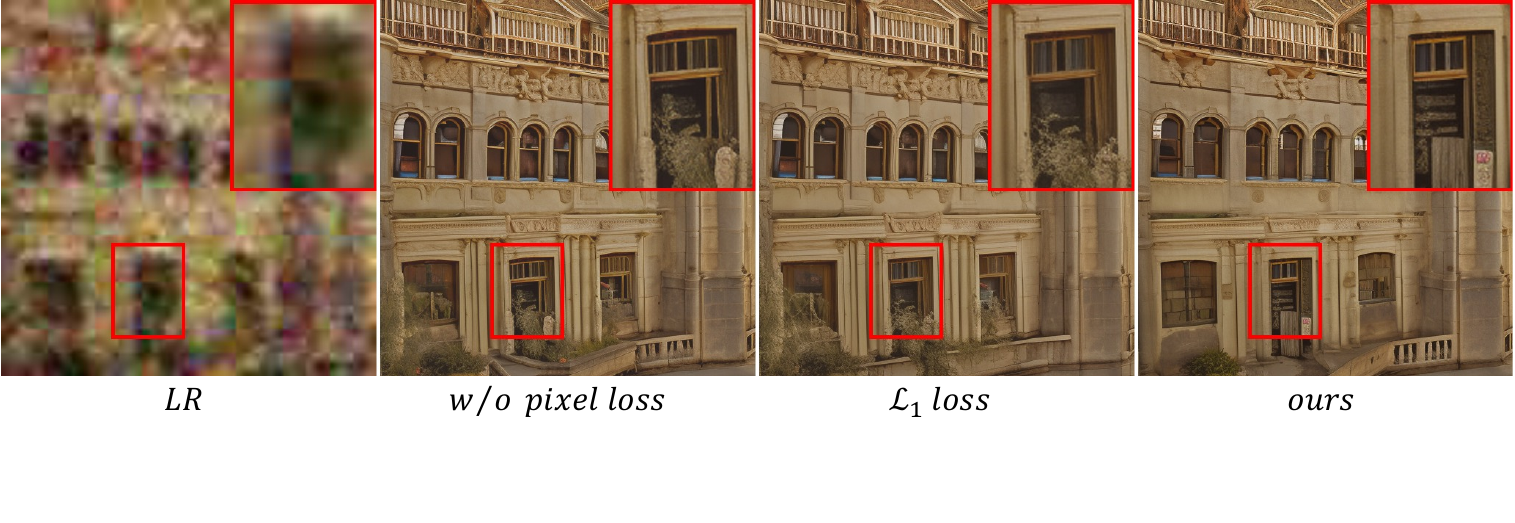}
  \caption{Effectiveness of Low-Frequency Constraints. ISR generates images with more structural content under low-frequency constraints.}
  \label{fig:ablation_dwt}
\end{figure}

\noindent \textbf{Effectiveness of Low-Frequency Constraints.}
Without pixel-level constraints, the ISR diffusion model achieves significantly higher perceptual metrics and lower fidelity metrics, as demonstrated in Table~\ref{tab:ablation_pixel_loss}. However, this improvement leads to structural inconsistencies in SR images, as illustrated in Figure~\ref{fig:ablation_dwt}, where disordered structures such as weeds appear in door frames. When $\mathcal{L}_{1}$ supervision replaces $\mathcal{L}_{dwt_{ll}}$, the SR images exhibit noticeable texture smoothing. This is because excessive supervision of high-frequency information during the early stages (low $st$) reduces the strength of low-frequency supervision, which weakens the ISR model's ability to maintain structural integrity and generate high-frequency details.

\begin{table}[!h]
    \centering
    \resizebox{0.45\textwidth}{!}{%
    \begin{tabular}{c ccccc} \toprule
      loss

      &$\text{LPIPS}\downarrow$ 
      &$\text{MANIQA}\uparrow$ 
      &$\text{MUSIQ}\uparrow$ 
      &$\text{CLIPIQA}\uparrow$ 
      &$\text{Aesthetic}\uparrow$ 
      
  \\ \midrule 

    w/o pixel loss  & 0.3382 & 0.6095 & 70.77 & 0.8153 & 5.2505	\\
    $\mathcal{L}_{1}$ & 0.3390 & 0.5828 & 69.73 & 0.7922 & 5.2418 \\
     \midrule 
    $\mathcal{L}_{dwt_{ll}}$  & 0.3369 & 0.5954 & 69.97 & 0.7944 & 5.2683 \\ \bottomrule
    \end{tabular}}
    \caption{{Ablations of Low-Frequency Constraints.}}
    \label{tab:ablation_pixel_loss}
\end{table}

\section{Conclusion}
\label{sec:cls}

In this study, we introduce reward feedback learning into ISR diffusion models by proposing a timestep-aware strategy. Specifically, during the initial denoising steps, we apply low-frequency information constraints to maintain the structural integrity of SR images. In the later denoising steps, we incorporate reward feedback learning to incentivize ISR models to generate SR images with improved perceptual and aesthetic quality. Extensive objective and subjective experiments validate that our method significantly improves the super-resolution performance of ISR diffusion models. We believe that reward feedback learning can become an important step in improving ISR diffusion models. While our method can fine-tune ISR to enhance performance, a limitation of our approach is that it relies on the generative quality of pre-trained SD models, which limits the maximum achievable fine-tuning performance. Moreover, the reward model used in our work, although commonly employed for image quality and aesthetic evaluation in academia, lacks robustness when confronted with larger-scale real-world data and diffusion-generated data.  
In future research, we plan to incorporate reward feedback learning into the training of ISR diffusion models from scratch and to develop more robust image quality evaluation models to guide ISR diffusion models.

{
    \small
    \bibliographystyle{ieeenat_fullname}
    \bibliography{main}
}

\end{document}